\documentclass[runningheads]{llncs}
\usepackage{graphicx}
\usepackage{amsmath,amssymb} 
\usepackage{color}
\usepackage[width=122mm,left=12mm,paperwidth=146mm,height=193mm,top=12mm,paperheight=217mm]{geometry}

\usepackage{expl3}
\ExplSyntaxOn
\newcommand\latinabbrev[1]{
  \peek_meaning:NTF . {
    #1\@}%
  { \peek_catcode:NTF a {
      #1.\@ }%
    {#1.\@}}}
\ExplSyntaxOff
\usepackage{nicefrac}

\def\eg{\latinabbrev{e.g}}
\def\etal{\emph{\latinabbrev{et al}}}

\usepackage{subfigure}

\begin{document}
\pagestyle{headings}
\mainmatter

\title{Connectionist Temporal Modeling for Weakly Supervised Action Labeling} 

\titlerunning{Connectionist Temporal Modeling for Weakly Supervised Action Labeling}

\authorrunning{De-An Huang, Li Fei-Fei, Juan Carlos Niebles}

\author{De-An Huang, Li Fei-Fei, Juan Carlos Niebles}


\institute{Computer Science Department, Stanford University \\ \email{\{dahuang,feifeili,jniebles\}@cs.stanford.edu}
}

\maketitle

\begin{abstract}
We propose a weakly-supervised framework for action labeling in video, where only the order of occurring actions is required during training time. The key challenge is that the per-frame alignments between the input (video) and label (action) sequences are unknown during training. We address this by introducing the Extended Connectionist Temporal Classification (ECTC) framework to efficiently evaluate all possible alignments via dynamic programming and explicitly enforce their consistency with frame-to-frame visual similarities. This protects the model from distractions of visually inconsistent or degenerated alignments without the need of temporal supervision. We further extend our framework to the semi-supervised case when a few frames are sparsely annotated in a video. With less than 1\% of labeled frames per video, our method is able to outperform existing semi-supervised approaches and achieve comparable performance to that of fully supervised approaches.
\end{abstract}

\section{Introduction}
With the rising popularity of video sharing sites like YouTube, there is a large amount of visual data uploaded to the Internet. This has stimulated recent developments of large-scale action understanding in videos \cite{THUMOS15,heilbron2015activitynet,KarpathyCVPR14}. Supervised learning methods can be effective in this case, but fully annotating actions in videos at large scale is costly in practice. An alternative is to develop methods that require weak supervision, which may be automatically extracted from movie scripts~\cite{bojanowski2013finding,bojanowski2014weakly,duchenne2009automatic} or instructional videos~\cite{alayrac2015learning,bojanowski2015weakly,yu2014instructional} at a lower cost.

In this work, we address the problem of weakly-supervised action labeling in videos. In this setting, only incomplete temporal localization of actions is available during training, and the goal is to train models that can be applied in new videos to annotate each frame with an action label. This is challenging as the algorithm must reason not only about whether an action occurs in a video, but also about its exact temporal location. Our setting contrasts with most existing works~\cite{kuehne2014language,messing2009activity,rohrbach2012database,yeung2015every} for action labeling that require fully annotated videos with accurate per frame action labels for training. Here, we aim at achieving comparable temporal action localization \emph{without} temporal supervision in training.

\begin{figure}[tb]
\centering
   \includegraphics[width=0.8\linewidth]{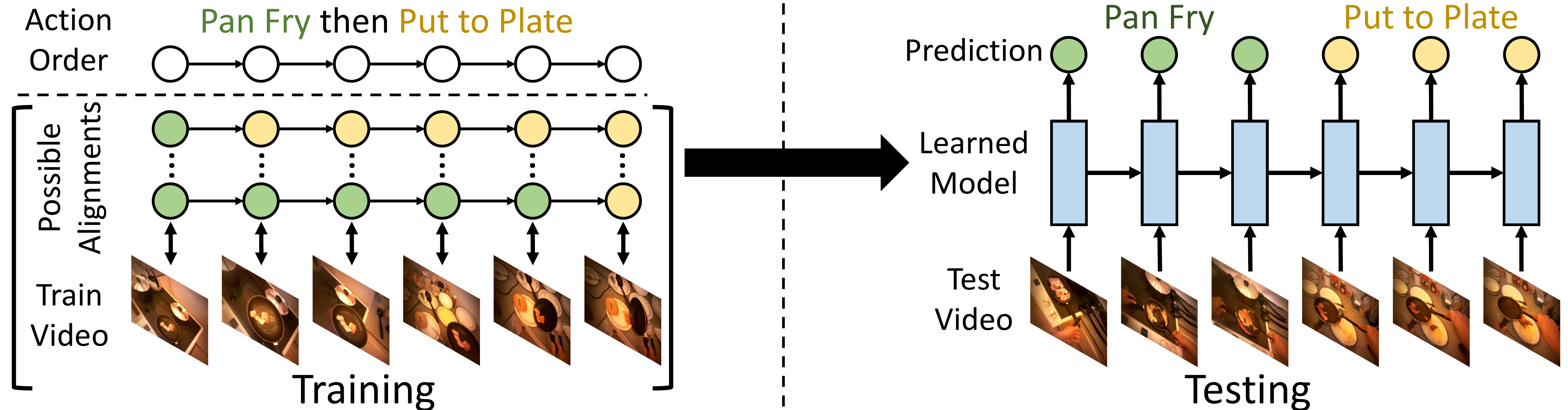}
   \caption{We tackle the problem of weakly supervised action labeling where
   only the order of the occurring actions is given during training (left).
   We train a temporal model by maximizing the probability of all possible
   frame-to-label alignments. At testing time (right), no annotation is given. As our learned model already encodes the temporal structure of videos, it predicts the correct actions without further information.}
\label{fig:fig1}
\end{figure}

The setting of our work is illustrated in Figure~\ref{fig:fig1}. During training, only the order of the occurring actions is given, and the goal is to apply the learned model to unseen test videos. As no temporal localization is provided during training, the first challenge of our task is that there is a large number of possible alignments (or correspondences) between action labels and video frames, and it is infeasible to naively search through all of these alignments. We address this challenge by first introducing Connectionist Temporal Classification (CTC)~\cite{graves2006connectionist}, originally designed for speech recognition, to our video understanding task. CTC efficiently evaluates all of the possible alignments using dynamic programming.

Directly applying the original CTC framework to our weakly-supervised action labeling could not fully address the challenge of a large space of possible frame to action label alignments. Note that the duration of an action could be hundreds of frames, which is much longer than the duration of phonetic states in speech recognition. As a result, we are required to align videos of thousands of frames to only dozens of actions. This poses a unique challenge in comparison to speech, as our space of possible alignments is much larger and contains degenerated alignments that can deteriorate performance. We address this challenge by proposing the Extended Connectionist Temporal Classification (ECTC) framework, which explicitly enforces the alignments to be consistent with frame-to-frame visual similarities. The incorporation of similarity allows us to (1) explicitly encourage the model to output visually consistent alignments instead of fitting to the giant space of all possible alignments (2) down-weight degenerated paths that are visually inconsistent. In addition, we extend the forward-backward algorithm of~\cite{graves2006connectionist} to incorporate visual similarity, which allows us to efficiently evaluate all of the possible alignments while explicitly enforcing their consistency with frame-to-frame similarities at the same time.

While our main focus is the weakly supervised setting, we also show how to extend our approach to incorporate supervision beyond action ordering. To this end, we introduce the \emph{frame-level} semi-supervised setting, where action labels are temporally localized in a few annotated video frames. This supervision could be  extracted from movie scripts~\cite{duchenne2009automatic,laptev2008learning} or by asking annotators to label actions for a small number of frames in the video, which is less costly than precisely annotating temporal boundaries of all actions. We model such supervision as a frame to label alignment constraints and naturally incorporate it in our ECTC framework to efficiently prune out inconsistent alignments. This significantly reduces the alignment space and boosts the performance of our approach. 

The main contributions of our work can be summarized as: (1) We first introduce CTC to our video understanding task, as a way to efficiently evaluate all frame to action alignments. (2) We propose ECTC to explicitly enforce the consistency of alignments with visual similarities, which protects the model from distractions of visually inconsistent alignments without the need of temporal supervision. (3) We extend ECTC to incorporate \emph{frame-level} semi-supervision in a unified framework to significantly reduce the space of possible alignments. (4) We test our model on long and complex activity videos from the Breakfast Actions Dataset~\cite{kuehne2014language} and a subset of the Hollywood2 dataset~\cite{bojanowski2014weakly}, and show that our method achieves state-of-the-art performance with less than 1\% of supervision.

\section{Related Work}

As significant progress has been made on categorizing temporally trimmed video clips, recent research of human activity analysis is shifting towards a higher level understanding in real-world settings~\cite{gkioxari2015finding,lan2015action,pirsiavash2014parsing,sener2015unsupervised,vo2014stochastic,yeung2015end}. Two tasks of action labeling have been explored extensively. The first is video classification, where the goal is to categorize each video to a discrete action class. Challenging datasets including UCF101~\cite{UCF101}, HMDB51~\cite{kuehne2011hmdb}, Sports1M~\cite{KarpathyCVPR14}, and ActivityNet~\cite{heilbron2015activitynet} exemplify this. Deep neural networks trained directly from videos~\cite{donahue2014long,simonyan2014two} have shown promising results on this task~\cite{xu2015UTS}. The second is dense action labeling, where the goal is to label each frame with the occurring actions~\cite{kuehne2014language,lillo2014discriminative,messing2009activity,rohrbach2012database,yeung2015every}, and the fully annotated temporal boundaries of actions are given during training.

In this paper, we aim to achieve action labeling with a weaker level of supervision that is easier to obtain than accurately time-stamped action labels. A similar goal has been explored in video to action
alignment~\cite{bojanowski2015weakly,bojanowski2014weakly}. The closest to our work is the ordering constrained discriminative clustering (OCDC) approach~\cite{bojanowski2014weakly}, where the goal is to align video frames to an ordered list of actions. Using the ordering constraint, OCDC extends previous work~\cite{duchenne2009automatic} to deal with multiple actions in a video. As their focus is on video to action alignment, their method can assume that the ordering of actions is available both at training and testing. Our approach aims at a more general scenario, where the learned model is applied to unseen test videos that come without information about the actions appearing in the video. When applied to this more general  scenario, OCDC is equivalent to a frame-by-frame action classifier that was implicitly learned during the training alignment. Therefore, OCDC does not fully exploit temporal information at test time, since it does not  encapsulate the temporal relationships provided by the ordering supervision. This may limit its applicability to temporally structured complex activities. On the other hand, our temporal modeling exploits the temporal structure of actions in videos, such as the transitions between actions, by capturing them during training and leveraging at test time.

Our work is also related to recent progress on using instructional videos or movie scripts~\cite{alayrac2015learning,bojanowski2015weakly,duchenne2009automatic,malmaud2015s,ramanathan2014linking,sener2015unsupervised,zhu2015aligning} as supervision for video parsing. These approaches also tackle the case when some text is available for alignment at testing time, and focus more on the natural language processing side of understanding the text in the instructions or the scripts. In this paper, we focus on training a temporal model that is applicable to unseen test videos that come without associated text. Our supervision could potentially be obtained with some of these text processing approaches, but this is not the focus of our work.

Our goal of understanding the temporal structure of video is related to~\cite{fernando2015modeling,ramanathan2015learning,ryoo2009spatio,song2013action,tang2012learning,taylor2010convolutional}. In contrast to their goal of classifying the whole video to a single action, our goal is to utilize the temporal structure of videos to guide the training of an action labeling model that can predict the occurring action at every frame in the unseen test video.
Our use of visual similarities in the training is related to unsupervised video parsing~\cite{niebles2008unsupervised,pirsiavash2014parsing,wu2015watch}, where frames are grouped into segments based on visual or semantic cues. We integrate visual similarity with weak supervision as a soft guidance of the model and go beyond just grouping video frames.

The core of our model builds upon Recurrent Neural Networks (RNN), which have been proved effective for capturing the temporal dependencies in data, and have been applied to challenging computer vision tasks including image captioning~\cite{chen2015mind,donahue2014long,karpathy2015deep}, video description~\cite{donahue2014long,venugopalan2015sequence,yao2015video}, activity recognition~\cite{donahue2014long,ng2015beyond}, dense video labeling~\cite{yeung2015every}. However, in the above tasks, accurate temporal localization of actions is either ignored
or requires pre-segmented training data. Our ECTC framework enables learning recurrent temporal models with weak supervision, and we show empirically its effectiveness on the video action labeling task.

\section{Ordering Constrained Video Action Labeling}
\label{sec:order}

Our goal is to train a temporal model to assign action labels to every frame of unseen test videos. We use a Recurrent Neural Network (RNN) at the core of our approach, as it has been successfully applied to label actions in videos~\cite{yeung2015every,donahue2014long}. While RNNs have been generally trained with full supervision in previous work, we aim to train them with weak supervision in the form of an ordered list of occurring actions. We address this challenge by proposing the Extended Connectionist Temporal Classification (ECTC) framework that efficiently evaluates all possible frame to action alignments and weights them by their consistency with the visual similarity of consecutive frames. The use of visual similarities sets our approach apart from the direct application of CTC~\cite{graves2006connectionist} and alleviates the problem caused by visually inconsistent alignments. ECTC incorporates a frame dependent binary term on top of the original unary based model, and we show that this can be efficiently handled by our forward-backward algorithm.

\subsection{Extended Connectionist Temporal Classification}
\label{sec:ectc}

\begin{figure}[tb]
\centering
   \includegraphics[width=0.85\linewidth]{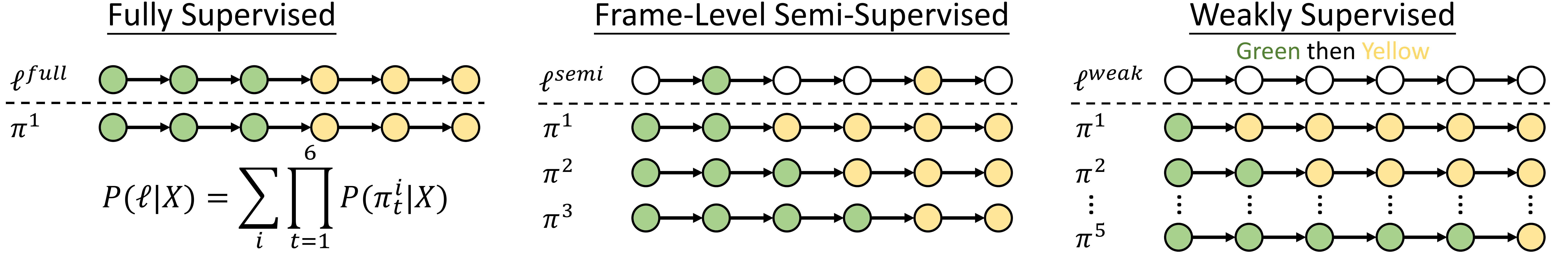}
   \caption{Comparison of different levels of supervision (first row).
   Blank circles indicate frames without annotated action. The probability of ${\ell}$ is given by the sum of the probabilities of all the paths $\pi^i$ that are consistent with it.}
\label{fig:comp-sup}
\end{figure}

The biggest challenge of our task is that only the order of the actions is given during training. Formally, given a training set consisting of video examples $X = [x_1, \cdots, x_{T}] \in \mathbb{R}^{d \times T}$ represented by $d$-dimensional features $x_t$ extracted from each of their $T$ frames, our goal is to infer the associated action labels $a = [a_1, \cdots, a_{T}] \in \mathcal{A}^{1\times T}$, where $\mathcal{A}$ is a fixed set of possible actions. Note that $a$ is not available for our training examples. Instead, the supervision we have for each video is the order of actions $\ell = \mathcal{B}(a)$, where $\mathcal{B}$ is the operator that removes the repeated labels. For example, $\mathcal{B}([b,b,c,c,c]) = [b,c]$. Our goal is to learn a temporal model using this supervision, and apply it to unseen test videos for which neither $\ell$ nor $a$ are available. We build our temporal models with an RNN at the core. Let $Y = [y_1, \cdots, y_T] \in \mathbb{R}^{A \times T}$ be the RNN output at each frame, where $A = |\mathcal{A}|$ is the number of possible actions. We normalize the output vectors $y_t$ using a softmax to get $z^k_t = P(k,t|X) = {e^{y_t^k}}/{\sum_{k'} e^{y^{k'}_t}}$, which can be interpreted as the probability of emitting action $k$ at time $t$.

In the original CTC formulation~\cite{graves2006connectionist}, the conditional independence assumption states that the probability of a label sequence $\pi = [\pi_1, \cdots, \pi_T]$ is:
\begin{equation}
\label{eq:ctc}
\small
P(\pi|X) = \prod_{t=1}^T z_t^{\pi_t},
\end{equation}
which corresponds to the stepwise product of $z_t^{\pi_t}$ at each frame. Note that we distinguish a \textit{path} $\pi$ that indicates per-frame label information from the \textit{label sequence} $\ell = \mathcal{B}(\pi)$ which only contains the ordering of actions and no precise temporal localization of labels. Label sequence $\ell$ is computed from path $\pi$ by $\mathcal{B}(\pi)$, which removes all the consecutive label repetitions.
We can compute the probability of emitting a label sequence $\ell$, by summing the probability of all paths $\pi$ that can be reduced to $\ell$ using the operator $\mathcal{B}$:
\begin{equation}
\label{eq:ctc_sum}
\small
P(\ell|X) = \sum_{\{\pi | \mathcal{B}(\pi) =\ell\}} P(\pi|X).
\end{equation}
Given the label sequence $\ell$ for each training video $X$, 
model learning is formulated as minimizing
$\mathcal{L}(\ell, X) = -\log P(\ell|X)$, the negative log likelihood of emitting $\ell$. The intuition is that, because we do not have the exact temporal location of a label, we sum over all the frame to label alignments that are consistent with $\ell$~\cite{graves2014towards}. 
One drawback of this original CTC formulation in Eq.\@(\ref{eq:ctc}) is that it does not take into account the fact that consecutive frames in the video are highly correlated, especially visually similar ones. This is important as the sum in Eq.\@(\ref{eq:ctc_sum}) might thus include label paths $\pi$ that are visually inconsistent with the video contents and thus deteriorate the performance. In the following, we discuss how our ECTC uses visual similarity to reweight the possible paths.

\begin{figure}[tb]
\centering
   \includegraphics[width=0.85\linewidth]{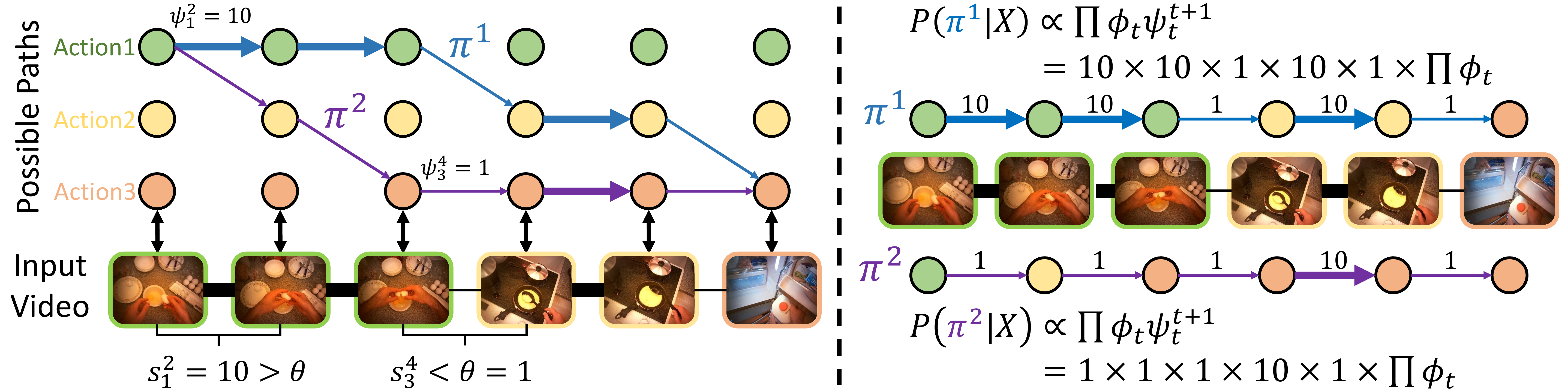}
   \caption{Our ECTC framework uses the binary term $\psi_t^{t+1}$ to re-weight paths. In this example, an input video has 6 frames and 3 annotated actions. Thicker connections between frames indicate higher similarity. In ECTC, $\pi^1$ has higher weight than $\pi^2$ since it stays in the same action for similar frames. In the example, $\pi^1$ actually matches the ground truth actions. In contrast, both paths are weighted equally in CTC.
   }
\label{fig:ectc}
\end{figure}

We introduce the Extended CTC framework to address such limitations. To illustrate our framework, assume that $z^a_t = z^b_t$ for all $t$ in a short clip of visually similar frames. In this example, the probability of the path $[a,b,a,b]$ will be the same as $[a,a,b,b]$ using Eq.\@(\ref{eq:ctc}). Clearly the latter path should be more probable, as action labels are usually smooth and stay the same for visually similar frames. Such intuition, however, is not compatible with Eq.\@(\ref{eq:ctc}). While our RNN could implicitly encode such pattern from training observations, we reformulate Eq.\@(\ref{eq:ctc}) to explicitly enforce the incorporation of visual similarity between consecutive frames by rewarding visually consistent paths:

\begin{equation}
\label{eq:our_path}
\small
P(\pi|X) \propto \prod_{t=1}^T \phi_t \psi_t^{t+1}, \quad \phi_t =z_t^{\pi_t}, \quad
\psi_t^{t+1} = \begin{cases}
\max(\theta, s_t^{t+1}) & \pi_t = \pi_{t+1}\\
\theta & \pi_t \neq \pi_{t+1}
\end{cases}.
\end{equation}
The path probability now includes both a unary term $\phi_t$ and a binary term $\psi_t^{t+1}$. The unary term is defined as $z_t^{\pi_t}$ and represents the original formulation. We introduce the binary term $\psi_{t}^{t+1}$ to explicitly capture the correlation between consecutive frames, 
where $\theta$ is a predefined minimum similarity, 
and $s_t^{t+1} = \text{sim}(x_t, x_{t+1})$ is the similarity between frames. When $\pi_t = \pi_{t+1}$ and $s_t^{t+1} > \theta$ (the two frames are similar), $\psi_{t}^{t+1} = s_t^{t+1}$ can be seen as a reward for staying at the same action. Effectively, our binary term explicitly rewards the paths that have the same action for visually similar frames, which further encourages the model to generate visually consistent action labels. On the other hand, frames with low similarity are not penalized for having the same action. When $\pi_t = \pi_{t+1}$ and $s_t^{t+1} < \theta$ (low similarity), $\psi_{t}^{t+1} = \theta$ is simply the same for all and has no effect on the path probability after normalization. Consider an extreme case when $s_t^{t+1} = \infty $. This effectively imposes the constraint that $\pi_t = \pi_{t+1}$, as the probability of paths with $\pi_t \neq \pi_{t+1}$ will be zero after normalization. As we will show in the experiment, our explicit modeling of the frame-to-frame correlation with the binary term plays an important role to the success of our model, as it allows us to avoid visually inconsistent and trivial paths in our task. Figure~\ref{fig:ectc} shows an example of how our ECTC reweights the paths using visual consistency.

\subsection{ECTC Forward-Backward Algorithm}
\label{sec:for-back}

At first sight, the summation in Eq.\@(\ref{eq:ctc_sum}) seems problematic, as the number of paths grows exponentially with the length of the input sequence. This is further complicated by the fact that our formulation in Eq.\@(\ref{eq:our_path}) involves a binary term $\psi_t^{t+1}$ that depends on both frame $t$ and $t+1$. We address this by proposing the ECTC forward-backward algorithm that extends the approach in~\cite{graves2006connectionist} and naturally incorporates the visual similarity function in a unified framework. We will show how the proposed algorithm is still able to efficiently evaluate all of the possible paths using dynamic programming despite the introduction of the binary term in Eq.\@(\ref{eq:our_path}) to explicitly capture the correlation between consecutive frames. We define our \textit{forward variable} as
{\small
\begin{align}
 \label{eq:forward}
\alpha(s,t) &= \sum_{\{ \pi_{1:t} | \mathcal{B}(\pi_{1:t})=\ell_{1:s}\}} P(\pi_{1:t}|X) \\
\label{eq:for-decomp}
&\propto \sum_{\{ \pi_{1:t} | \mathcal{B}(\pi_{1:t})=\ell_{1:s}\}} \prod_{t'=1}^t \Psi_{t'}^{\pi_{t'}}z_{t'}^{\pi_{t'}},
\quad \Psi_t^{k} = \begin{cases}
\max(\theta, s_{t-1}^{t}) & k = \pi_{t-1}\\
\theta & k \neq \pi_{t-1}
\end{cases},
\end{align}
}%
which corresponds to the sum of probabilities of paths with length $t$ $\pi_{1:t}=[\pi_1,\cdots,\pi_t]$ that satisfy $\mathcal{B}(\pi_{1:t})=\ell_{1:s}$, where $\ell_{1:s}$ is the first $s$ elements of the label sequence $\ell$.
We also introduce a new variable $\Psi_t^{k}$ for explicitly modeling the dependence between consecutive frames and encourage the model to output visually consistent path. This makes the the original CTC forward-backward algorithm not directly applicable to our formulation. 
By deriving all $\pi_{1:t}$ that satisfy $\mathcal{B}(\pi_{1:t})=\ell_{1:s}$ from $\pi_{1:t-1}$, the forward recursion is formulated as:
\begin{equation}
\label{eq:forward_update}
\small
\alpha(s,t) = \hat{z}^{\pi_t}_t \alpha(s, t-1) + \tilde{z}^{\pi_t}_t \alpha(s-1, t-1),
\end{equation}
where
{\small
\begin{align}
 \hat{z}^{\pi_t}_t &= \frac{\Psi^{\pi_t}_t z_{t}^{\pi_t}}{\sum_{k=1}^A \Psi_{t}^{k}z_{t}^{k}} = \frac{\max(\theta, s_{t-1}^t)z_{t}^{\pi_t}}{\max(\theta, s_{t-1}^t)z_{t}^{\pi_t} + \theta(1-z_{t}^{\pi_t})} \\
 \label{eq:z2}
 \tilde{z}^{\pi_t}_t &=  \frac{\Psi^{\pi_t}_t z_{t}^{\pi_t}}{\sum_{k=1}^A \Psi_{t}^{k}z_{t}^{k}}
  = \frac{\theta z_t^{\pi_t}}{\max(\theta, s_{t-1}^t)z_{t}^{\pi_{t-1}} + \theta(1-z_{t}^{\pi_{t-1}})}.
\end{align}
}%
The key difference between our algorithm and that of~\cite{graves2006connectionist} is the renormalization of $z^k_t$ using frame similarity $\Psi_t^k$, which in turn gives the renormalized $\hat{z}^{\pi_t}_t$ and $\tilde{z}^{\pi_t}_t$. This efficiently incorporates visual similarity in the dynamic programming framework and encourages the model towards visually consistent paths. The first term in Eq.\@(\ref{eq:forward_update}) corresponds to the case when $\pi_t = \pi_{t-1}$. Based on the definition, we have $\Psi_t^{\pi_t} = \Psi_t^{\pi_{t-1}} = \max(\theta,s_{t-1}^t)$. Intuitively, this reweighting using $\Psi_t^k$ will reward and raise $z^{\pi_t}_t$ to $\hat{z}^{\pi_t}_t$ for having the same action label for similar consecutive frames. On the other hand, the second term in Eq.\@(\ref{eq:forward_update}) is for the case when $\pi_t \neq \pi_{t-1}$, and thus $\Psi_t^{\pi_t} = \theta$. In this case, the probability is taken from $\tilde{z}^{\pi_t}_t$ to reward $\tilde{z}^{\pi_{t-1}}_t$, and thus $\tilde{z}^{\pi_t}_t$ will be smaller than $z^{\pi_t}_t$.

The \textit{backward variable} is similarly defined as:
\begin{equation}
\small
\beta(s,t) = \sum_{\{ \pi_{t:T} | \mathcal{B}(\pi_{t:T})=\ell_{s:S}\}} P(\pi_{t:T}|X)
\propto \sum_{\{ \pi_{t:T} | \mathcal{B}(\pi_{t:T})=\ell_{s:S}\}} \prod_{t'=t}^T \tilde{\Psi}_{t'}^{\pi_{t'}}z_{t'}^{\pi_{t'}},
\end{equation}
the sum of the probability of all paths starting at $t$ that will complete $\ell$ when appending from $t+1$ to any path of $\alpha(s,t)$. We also introduce $\tilde{\Psi}_t^{k}$ in the same way as $\Psi_t^{k}$, but by decomposing Eq.\@(\ref{eq:our_path}) backward rather than forward. The backward recursion to compute $\beta(s,t)$ can be derived similarly to the forward recursion in Eq.\@(\ref{eq:forward_update}), but by deriving $\pi_{t:T}$ from $\pi_{t+1:T}$. 
Based on the definition of forward and backward variables, we have
$P(\ell|X) = \sum_{s=1}^S \frac{\alpha(s,t)\beta(s,t)}{z_t^{\ell_s}}$.

\subsubsection{Optimization.}
With this forward-backward algorithm, we are able to compute the gradient of the loss function $\mathcal{L}(\ell, X)$ w.r.t. the recurrent neural network output $y_t^k$, the response of label $k$ at time $t$. The gradient is given by:
\begin{equation}
\small
\frac{\partial \mathcal{L}(\ell, X)}{\partial y_t^k} = z_t^k - \frac{1}{P(\ell|X)}\sum_{s:\ell_s=k} \frac{\alpha(s,t)\beta(s,t)}{z_t^{\ell_s}},
\end{equation}
where the first term is the softmax output. The second term can be seen as the softmax target.
The second term can be intuitively interpreted as $P(\pi_t = k | \mathcal{B}(\pi)=\ell, X)$, which is the probability of choosing action $k$ at time $t$ for paths that are consistent with the sequence label $\ell$ (reweighted by $\psi_t^{t+1}$). The recurrent neural network can then be optimized through back propagation~\cite{rumelhart1988learning}.

\section{Extension to Frame-level Semi-Supervised Learning}

\begin{figure}[tb]
\centering
   \includegraphics[width=1\linewidth]{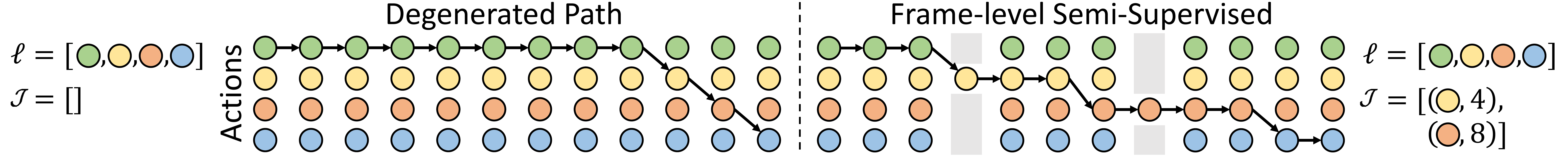}
   \caption{Example of a degenerated path and a semi-supervised path. On the right, gray blocks constrain the path to be consistent with the two supervised frames. This significantly reduces the space of possible paths and prevents  degenerated paths.}
\label{fig:cons-path}
\end{figure}

When only the ordering supervision is available, all of the paths $\pi$ that are consistent with $\ell$ are considered in Eq.\@(\ref{eq:ctc_sum}). A quick observation, however, shows that some undesirable or degenerate paths shown in Figure~\ref{fig:cons-path} are also considered in the summation. This challenge is unique to our task as the length of the label sequence $\ell$ is usually much shorter than the number of frames, which is not the case in speech recognition. We have shown how our ECTC can be used in this case as soft constraints to down-weight such visually inconsistent paths and reward the ones that have consistent labels for visually similar frames. Nevertheless, when supervision beyond ordering is available, we can derive harder constraints for the paths and effectively remove undesirable paths from the summation.

In this section, we show that sparse temporal supervision can also be naturally incorporated in our framework and efficiently prune out the degenerated paths. We introduce the \emph{frame-level} semi-supervised setting, where only a few frames in the video are annotated with the ground truth action. 
Such supervision could be automatically extracted from movie scripts~\cite{duchenne2009automatic,laptev2008learning} or by annotating a few frames of the video, which is much easier than finding the exact temporal boundaries of all the actions.
Formally, the supervision we consider is a list of frames with the corresponding action labels:
$
\mathcal{J} = [ (a_1, t_1), \cdots, (a_m,t_m),\cdots (a_M,t_M)],
$
where each element of the list is a pair of frame index $t_m$ and the corresponding action label $a_m$. This can significantly reduce the number of possible paths when combined with the order of the actions. For example, assuming that we have $\mathcal{J} = [(a,2),(b,4)]$ and $\ell = [a,b]$ for a video of length 6, then there are only two possible paths ($[a,a,b,b,b,b]$ and $[a,a,a,b,b,b]$) that are consistent with the supervision. This not only significantly reduces the space of consistent paths, but also avoids undesirable paths like $[a,a,a,a,a,b]$. Figure~\ref{fig:cons-path} also shows an example of the effect of the frame-level semi-supervision.
This supervision can be naturally incorporated by extending the recursion in Eq.\@(\ref{eq:forward_update}) as:
\begin{equation}
\small
\alpha(s,t) = \begin{cases}
0,\quad \exists (a_m,t_m) \in \mathcal{J},\text{ s.t. } t = t_m \text{ but } s \neq a_m \\
\hat{z}^{\pi_t}_t \alpha(s, t-1) + \tilde{z}^{\pi_t}_t \alpha(s-1, t-1), \quad  \text{otherwise}
\end{cases},
\end{equation}
where an extra checking step is applied to ensure that the path is consistent with the given semi-supervision.
We will show that, with less than 1\% of frames being labeled, our approach can perform comparably to fully supervised model.

\section{Experiments}

We evaluate our model on two challenging tasks and datasets. The first is segmentation of cooking activity video in the Breakfast Actions Dataset~\cite{kuehne2014language}. The output action labeling divides the video into temporal segments of cooking steps. Because of the dependencies between temporally adjacent actions in cooking activities, the capacity of the model to handle temporal dynamics is especially important. The second task is action detection on videos in a subset of the Hollywood2 dataset~\cite{laptev2008learning}, with a setting introduced by~\cite{bojanowski2014weakly}. Our action labeling framework can be applied to action detection by considering an additional background label $\emptyset$ to indicate frames without actions of interest.  

\subsection{Implementation Details}

{\noindent \bf Network Architecture.} We use 1-layer Bidirectional LSTM (BLSTM)~\cite{graves2005framewise} with 256 hidden units for our approach. We cross-validate the learning rate and the weight decay. For the optimization, we use SGD with batch size 1. We clip gradients elementwise at 5 and scale gradients using RMSProp~\cite{RMSProp}.

{\noindent \bf Visual Similarity.} For our ECTC, we combine two types of visual similarity functions. The first is clustering of visually similar and temporally adjacent frames. We apply $k$-means clustering to frames in a way similar to SLIC~\cite{achanta2012slic} to over-segment the video. We initialize $\frac{T}{M}$ centers uniformly for a video, where $T$ is the video length, and $M$ is the average number of frames in a cluster.
We empirically pick $M=20$, which is much shorter than the average length of an action ($\sim$400 frames in the Breakfast Dataset) to conservatively over segment the video and avoid grouping frames that belong to different actions.
The resulting grouping is in the form of constraints such as 
$\pi_t = \pi_{t+1}$, which can be easily incorporated in our ECTC by setting $s_t^{t+1}$ to $\infty$. We thus set $s_t^{t+1}=\infty$ if the video frames $x_t$ and $x_{t+1}$ are in the same cluster and $s_t^{t+1}=0$ otherwise.
The second visual similarity function we consider is $s_t^{t+1}\propto \frac{x_t \cdot x_{t+1}}{|x_t||x_{t+1}|}$, the cosine similarity of the frames. This formulation will reward paths that assign visually similar frames to the same action and guide the search of alignment during the forward-backward algorithm. We combine the two similarity functions by setting $s_t^{t+1}$ to the cosine similarity at the boundary between clusters instead of $0$.

\subsection{Evaluating Complex Activity Segmentation}

In this task, the goal is to segment long activity videos into actions composing the activity. We follow~\cite{kuehne2014language} and define the \emph{action units} as the shorter atomic actions that compose the longer and more complex activity. For example, ``Take Cup'' and ``Pour Coffee'' are action units that compose the activity ``Make Coffee''.

{\noindent \bf Dataset.} We evaluate activity segmentation of our model on the Breakfast dataset~\cite{kuehne2014language}.
The videos of the dataset were recorded from 52 participants in 18 different kitchens conducting 10 distinct cooking activities.
This results in $\sim$77 hours of videos 
of preparing dishes such as fruit salad and scrambled eggs.

{\noindent \bf  Metrics.} We follow the metrics used in previous work~\cite{kuehne2014language} to evaluate the parsing and segmentation of action units. The first is \emph{frame accuracy}, the percentage of frames that are correctly labeled. The second is \emph{unit accuracy}. 
The output action units sequence is first aligned to the ground truth sequence by dynamic time warping (DTW) before the error rate is computed. 
For weakly supervised approaches, high frame accuracy is harder to achieve than high unit accuracy because it directly measures the quality of the temporal localization of actions.

{\noindent \bf  Features.}
We follow the feature extraction steps of~\cite{kuehne2015towards} and use them for all competing methods. First, the improved dense trajectory descriptor~\cite{wang2013action} is extracted and encoded by Fisher Vector with GMMs=64. L2 and power normalization, and PCA dimension reduction ($d=64$) are then applied.

{\noindent \bf Baselines.} We compare our method to three baselines. The first is per-frame Support Vector Machine ({\bf SVM})~\cite{CC01a} with RBF kernels.We are interested in how well discriminative classification can do on the video segmentation task without exploiting the temporal information in the videos. The second is Hidden Markov Model Toolkit ({\bf HTK}) used in previous work for this task~\cite{kuehne2014language,kuehne2015towards}.
The third is Order Constrained Discriminative Clustering ({\bf OCDC}) of Bojanowski \etal~\cite{bojanowski2014weakly}, which has been applied to align video frames with actions.

\begin{table}[b]
\caption{Ablation studies of our approach on the Breakfast dataset. Each component introduced in our approach gives an accuracy improvement. Our final ECTC model is able to outperform fully supervised baselines.}
\label{tab:BF}
\centering
\scriptsize
\begin{tabular}{|c|c|c||c|c|c|c|c|c|c|}
\hline
Supervision     & \multicolumn{2}{c||}{{\bf Fully Sup.}} & \multicolumn{6}{c|}{{\bf Weakly Sup.}} \\ \hline
Model      & 
\begin{tabular}[c]{@{}c@{}}SVM\\\cite{CC01a}\end{tabular} & 
\begin{tabular}[c]{@{}c@{}}HTK\\\cite{kuehne2014language}\end{tabular} & 
\begin{tabular}[c]{@{}c@{}}OCDC\\\cite{bojanowski2014weakly}\end{tabular} & 
\begin{tabular}[c]{@{}c@{}}Uniform\end{tabular} &
\begin{tabular}[c]{@{}c@{}}CTC\end{tabular} & 
\begin{tabular}[c]{@{}c@{}}ECTC\\(kmeans)\end{tabular} & 
\begin{tabular}[c]{@{}c@{}}ECTC\\(cosine)\end{tabular} & 
\begin{tabular}[c]{@{}c@{}}ECTC\\(Our full model)\end{tabular} \\ \hline
Frame Acc. & 15.8 & 19.7   & 8.9  & 22.6 & 21.8 & 24.5 & 22.5 & {\bf 27.7} \\ \hline
Unit Acc.  & 15.7 & 20.4   & 10.4 & 33.1 & 36.3 & 35.0 & {\bf 36.7} & 35.6 \\ \hline
\end{tabular}
\end{table}

\begin{figure}[t]
\centering
   \includegraphics[width=0.95\linewidth]{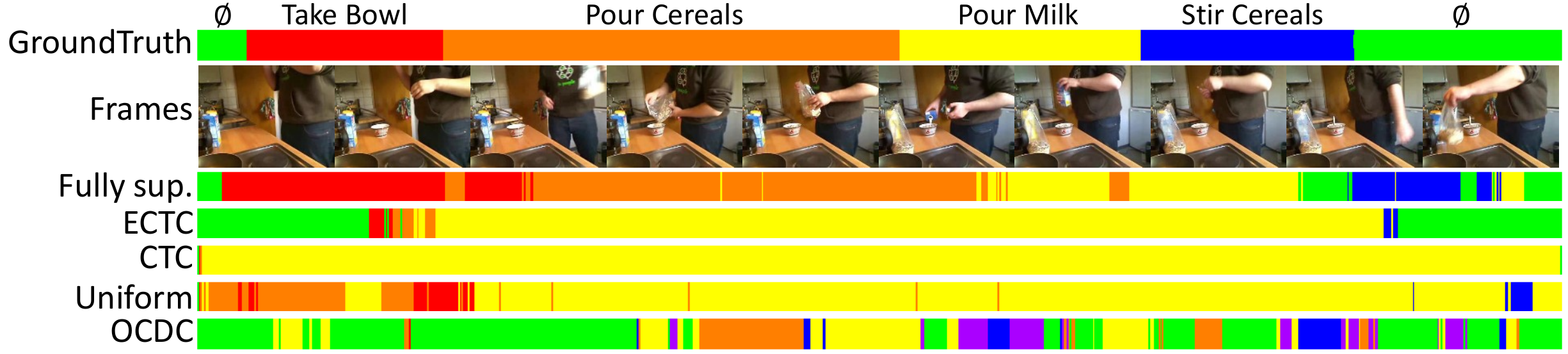}
   \caption{ Qualitative comparison of weakly supervised approaches in a testing video. Fully supervised results using BLSTM are also shown as reference (upper bound of our approach). Colors indicate different actions, and the horizontal axis is time. Per frame classification of OCDC is noisy and contains unrelated actions. The Uniform baseline produces the proper actions, but without alignment and ordering. CTC outputs a degenerated path in this case: while the order is correct, the sequence is dominated by a single action. Our ECTC has better localization and ordering of the actions since we incorporate visual similarity to prune out inconsistent and degenerated paths. 
   }
\label{fig:BF-vis}
\end{figure}

{\noindent \bf Ablation Studies.} First we analyze the effect of different components of our approach and compare to the baselines. The results are shown in Table~\ref{tab:BF}. 
The first variation of our model is ``{\bf Uniform}''. Instead of using our framework to evaluate all possible paths, the target of Uniform is a single path $\pi$ given by uniformly distributing the occurring actions among frames.
We also show the performance of direct application of {\bf CTC} to our task. Without explicitly imposing the alignments to be consistent with the visual similarity, CTC only has the effect of trading-off frame accuracy for unit accuracy when compared to the Uniform baseline. The reason is that the original CTC objective is actually directly optimizing the unit accuracy, but ignoring the frame accuracy as long as the output action order is correct.
The performance of our ECTC with only the clustering similarity is shown as ``{\bf ECTC (kmeans)}''. This efficiently rules out the paths that are inconsistent with the visual grouping and improve the alignment of actions to frames. Using only the cosine similarity with ECTC (``{\bf ECTC (cosine)}''), we are able to further improve the unit accuracy. Combining both similarities, the last column of Table~\ref{tab:BF} is our final ECTC model, which further improves the accuracy and outperforms fully supervised baselines. This verifies the advantage of using visual similarity in our ECTC to reward paths that are consistent with it. All variations of our temporal models outperform the linear classifier in OCDC on unseen test videos. Figure~\ref{fig:BF-vis} shows the qualitative results.

\begin{figure}[tb]
\centering
   \includegraphics[width=0.9\linewidth]{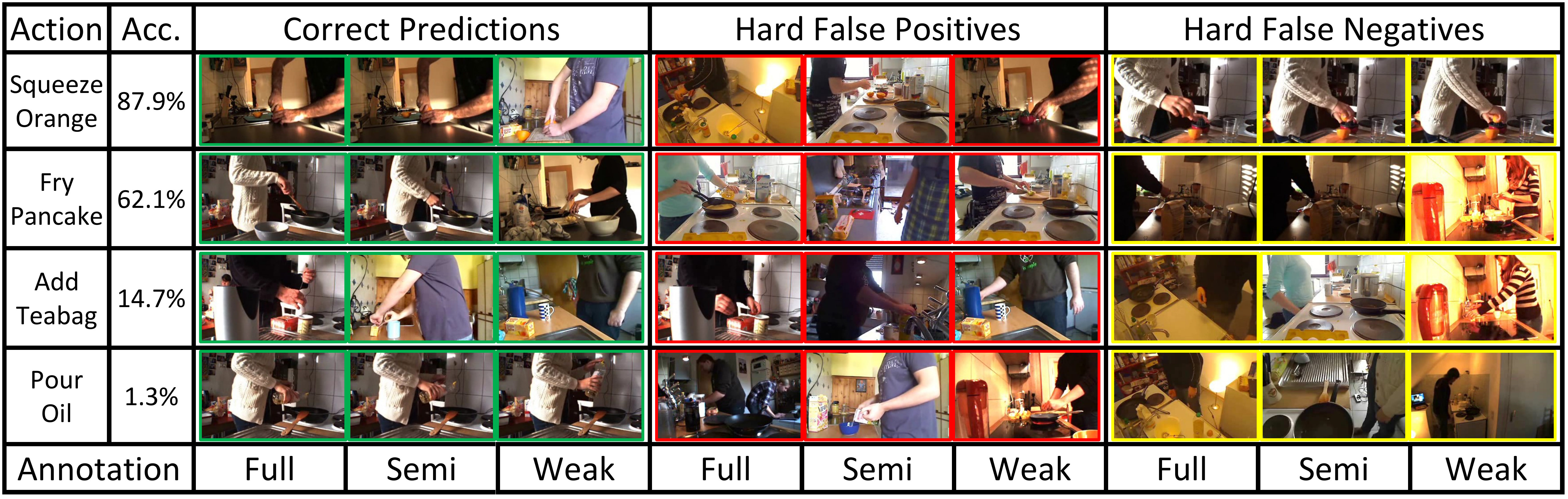}
   \vspace{-2mm}
   \caption{Example results for the two hardest and easiest actions. Correct Predictions illustrate the most confident correct frame predictions. Hard False Positives show incorrect predictions with high confidence. Our models can be confused by the appearance of objects in the scene (\eg, seeing the teabag box), or by similar motions (\eg, pouring milk instead of oil). Hard false negative show missing detections. We see challenges of viewpoint, illumination, and  ambiguities near the annotated boundary between actions. }
   \vspace{-1mm}
\label{fig:BF-TFPN}
\end{figure}

\begin{figure}[b]
\centering
   \includegraphics[width=0.59\linewidth]{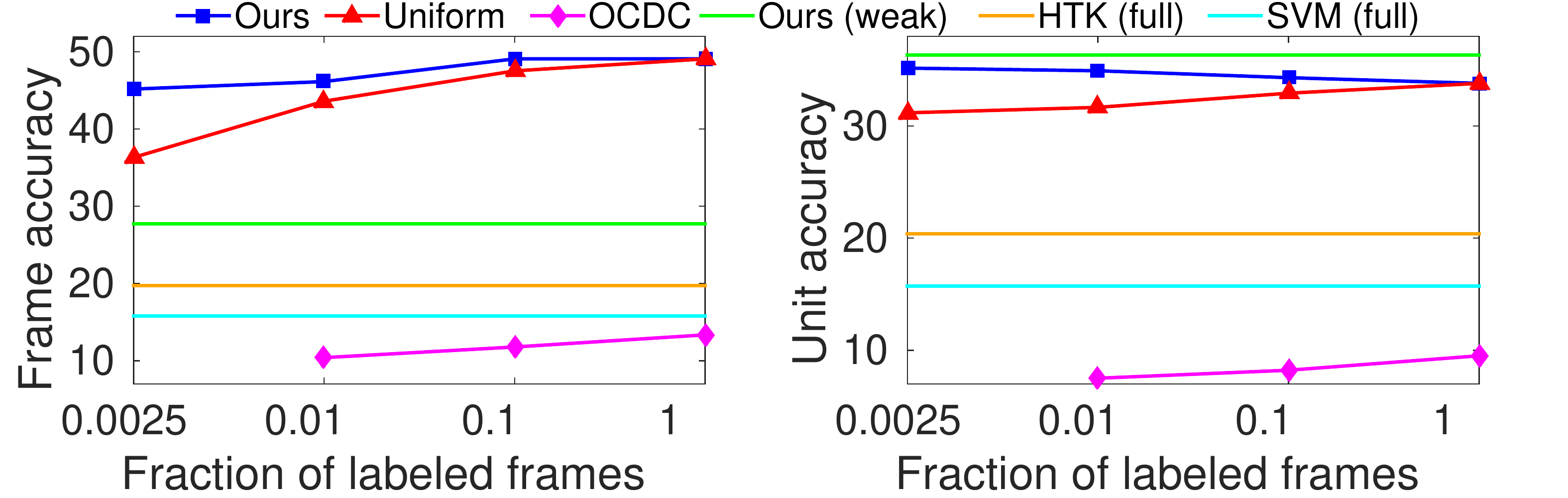}
   \includegraphics[width=0.38\linewidth]{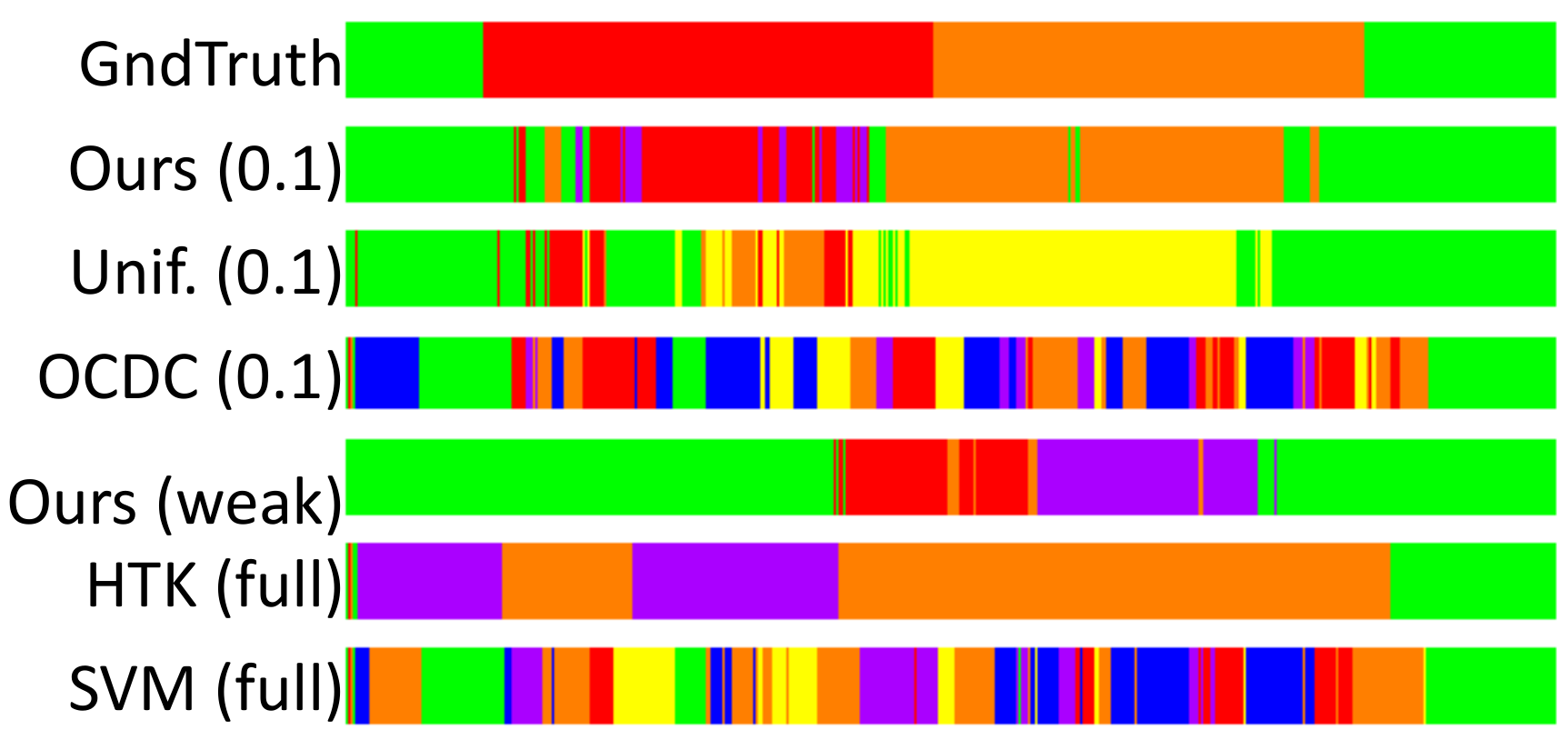}
   \caption{Frame and unit accuracy in the Breakfast dataset plotted against fraction of labeled data in the frame-level semi-supervised setting. Horizontal lines are either fully-supervised or weakly supervised methods. On the right, qualitative results for one video follow the convention of Figure~\ref{fig:BF-vis}.
   }
\label{fig:BF}
\end{figure}

{\noindent \bf Frame-level Semi-Supervision.} Next we study the effect of having more supervision with our model. The results are shown in Figure~\ref{fig:BF}. The $x$-axis shows the fraction of labeled frames for our frame-level semi-supervision in each video. 
The minimum supervision we use is when only a single frame is labeled for each occurring action in the video (fraction 0.0025). Fraction $1$ indicates our fully supervised performance.  
The annotation for the Uniform baseline in this case is equally distributed between two sparsely annotated frames. With our approach, the frame accuracy is dropping much slower than that of the Uniform baseline, since our approach is able to automatically and iteratively refine the alignment between action and frame during training. Our semi-supervised approach significantly outperforms OCDC with all fractions of labeled frames. The results of HTK, SVM, and our full ECTC are also plotted for reference. As noted earlier, our weakly supervised approach has the highest unit accuracy, as the CTC objective is directly optimizing it. This is consistent with the fact that lower fraction of labeled frames of our approach actually has higher unit accuracy. Another interesting observation is the gap between our weakly supervised model and semi-supervised model. While our weakly supervised model already outperforms several baselines with full supervision, it can be seen that giving only a single frame annotation as an anchor for each segment significantly reduces the space of possible alignments and  provides a strong cue to train our temporal model.
Figure~\ref{fig:BF-TFPN} shows results for different levels of supervision.

\begin{figure}[tb]
\centering
   \includegraphics[width=0.59\linewidth]{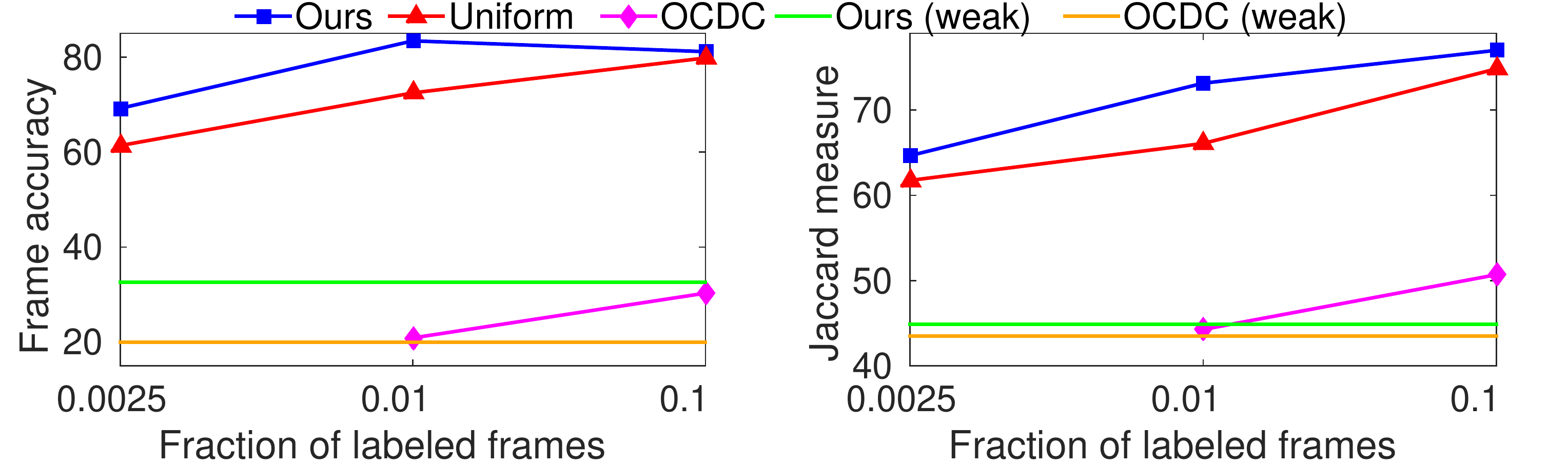}
   \includegraphics[width=0.38\linewidth]{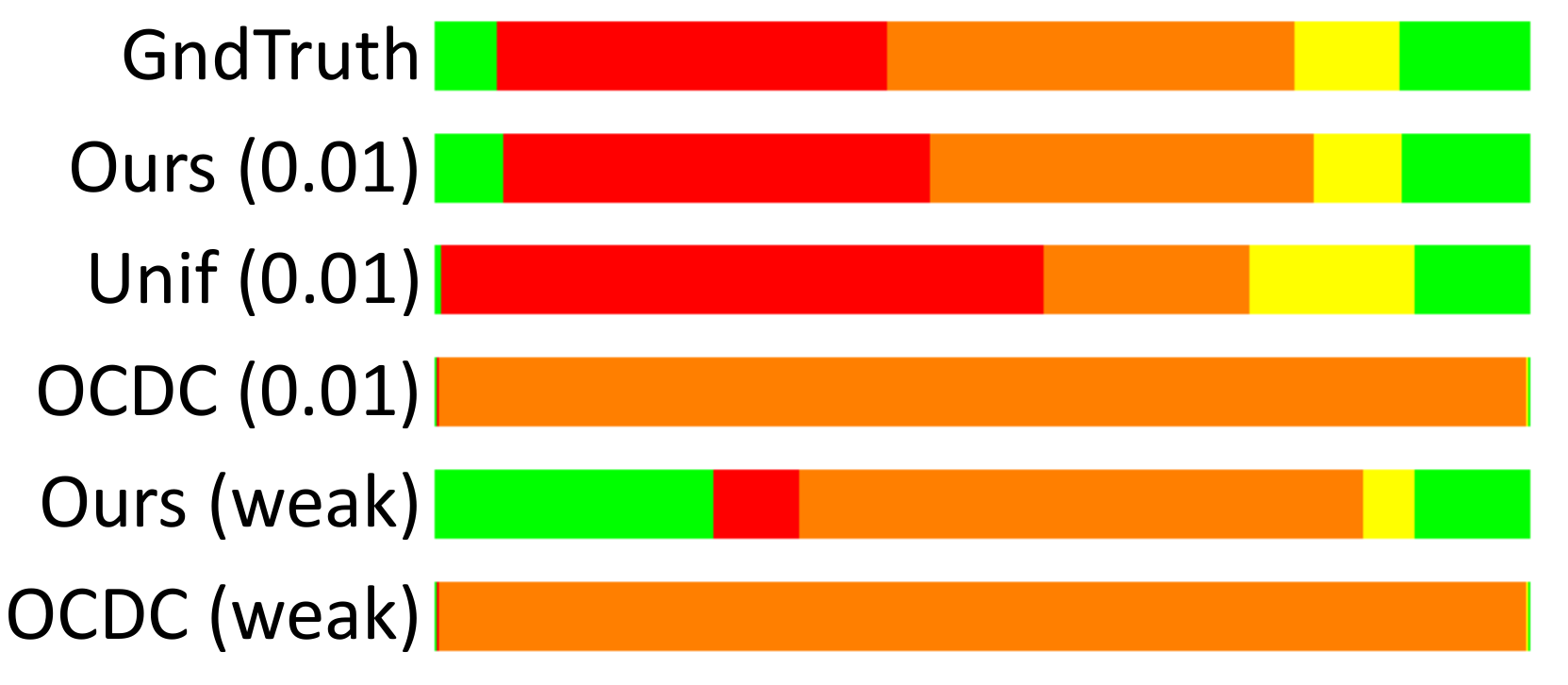}
   \caption{Frame accuracy, Jaccard measure, and qualitative alignment results on the training set of the Breakfast dataset. Our models also produce good alignments for the training set in addition to the ability to segment unseen test videos.}
\label{fig:BF-align}
\end{figure}

{\noindent \bf Training Set Alignment.}
While our framework aims at labeling unseen test videos when trained only with the ordering supervision, we also verify whether our action-frame alignment during training also outperforms the baselines. The frame accuracy and Jaccard measure are shown in Figure~\ref{fig:BF-align}. Jaccard measure is used to evaluate the alignment quality in~\cite{bojanowski2014weakly}. OCDC that is directly designed for the alignment problem indeed performs closer to our method in this scenario. 

\subsection{Evaluating Action Detection}

\begin{figure}[b]
\centering
   \includegraphics[width=0.64\linewidth]{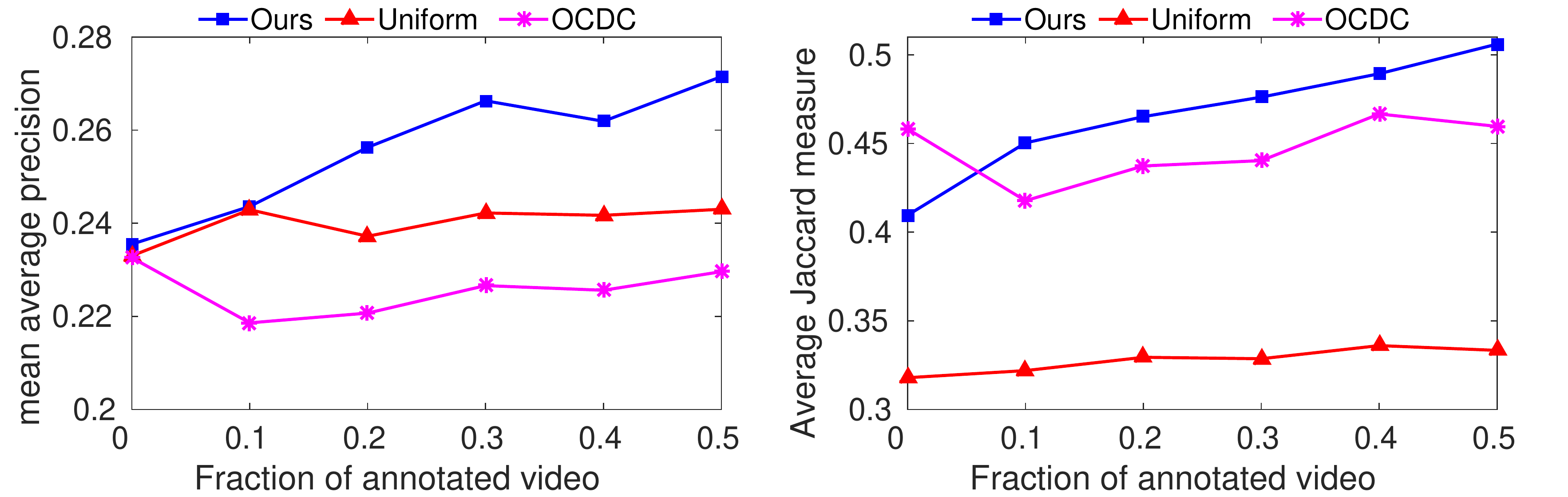}
   \includegraphics[width=0.32\linewidth]{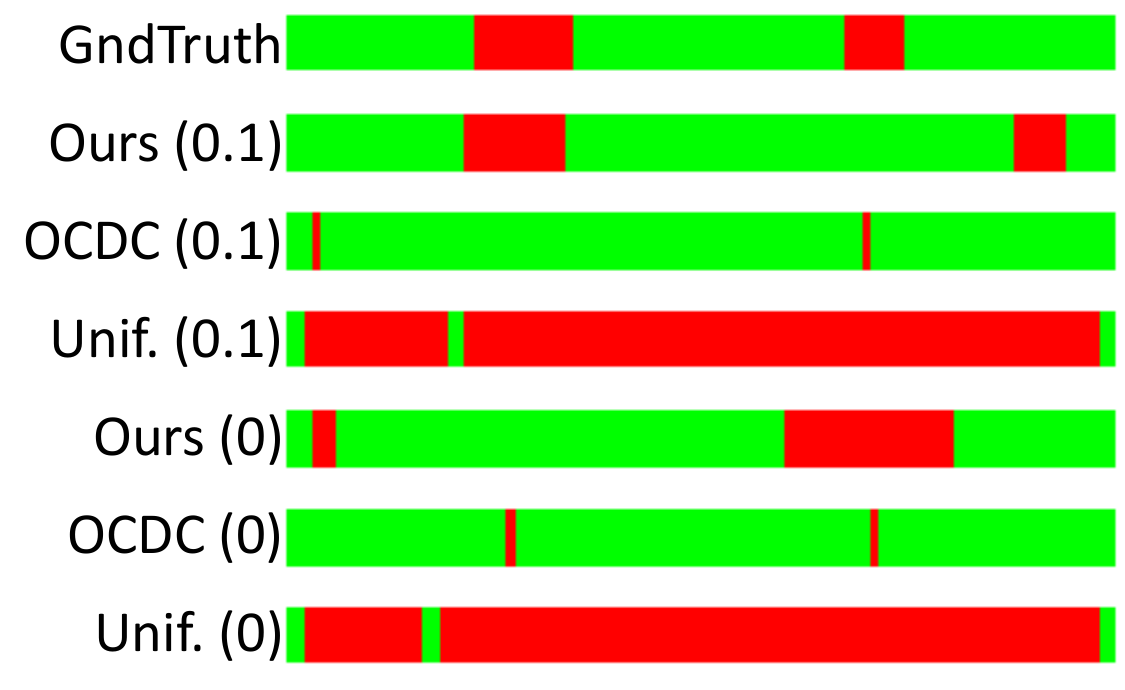}
\vspace{-2mm}
   \caption{Left plots mAP for action detection in unseen test videos. Middle plots the average Jaccard measure on the alignment evaluation set. Note that zero fraction of annotated video corresponds to the weakly-supervised setting, where all the videos in training set only have ordering supervision. Our approach consistently outperforms both baselines because of our temporal modeling and efficient evaluation of all possible alignments. On the right, we illustrate qualitative alignment results for all methods.}
\label{fig:AO}
\end{figure}

In this task, the goal is to localize actions in the video. This can be formulated as action labeling by introducing the background label $\emptyset$ to indicate frames without actions of interest. One practical challenge of this task is that the videos tend to be dominated by $\emptyset$. This requires the model to deal with unbalanced data and poses a different challenge than the temporal segmentation task.

{\noindent \bf Dataset and Metrics.} 
We evaluate action detection of our model on the dataset of Bojanowski \etal~\cite{bojanowski2014weakly}, which consists of clips taken from the 69 movies Hollywood2~\cite{laptev2008learning} dataset were extracted. The full time-stamped annotation of 16 actions (\eg ``Open Door'' and ``Stand Up'') are manually added. For metrics, we follow~\cite{bojanowski2014weakly} and use mean average precision for evaluating action detection and average Jaccard measure for evaluating the action alignment. 

{\noindent \bf Experimental Setup.} We use the extracted features from Bojanowski \etal~\cite{bojanowski2014weakly} for all the methods. All methods use the same random splitting of the dataset. As we follow the exact setup of~\cite{bojanowski2014weakly} for  evaluation, we would like to clarify that the semi-supervised here means \emph{video-level} semi-supervised setting, where a fraction of the videos in the \emph{supervised} set has full supervision, while the rest only has ordering as supervision. In this sense, the $0$ fraction corresponds exactly to our weakly supervised setting, where all the videos only have ordering supervision. This is different from the \emph{frame-level} semi-supervised setting we have discussed. All experiments are conducted over five random splits of the data.

\begin{figure}[tb]
\centering
   \includegraphics[width=0.85\linewidth]{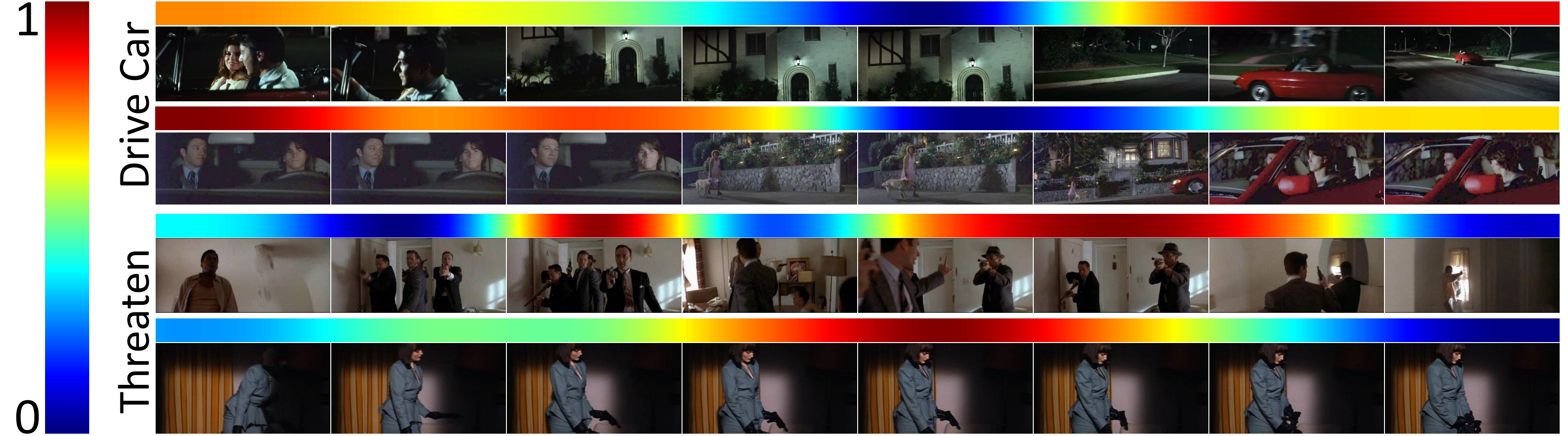}
   \caption{Our weakly-supervised action detection results. Color means the output probability of the target action. Our model accurately localizes actions of varied lengths.}
   \label{fig:AO-det}
\end{figure}

{\noindent \bf Detection Results.} The action detection results on the held-out testing set are shown in Figure~\ref{fig:AO} (left). While the occurring actions do not have a strong correlation, the results still demonstrate the importance of temporal modeling for better performance on held-out data. Both of our approaches outperform the OCDC baseline of Bojanowski \etal~\cite{bojanowski2014weakly} in this scenario. Figure~\ref{fig:AO-det} shows the qualitative results of our weakly-supervised action detection model.

{\noindent \bf Alignment Results.} The action alignment result on the evaluation set is shown in Figure~\ref{fig:AO} (middle). The uniform baseline performs the worst in this scenario, as there is no refinement of the alignment.
On the other hand, our ECTC incorporates visual similarity and efficiently evaluates all possible alignments. This allows it to perform the best even for the alignment problem.

\section{Conclusions}

We have presented ECTC, a novel approach for learning temporal models of actions in a weakly supervised setting. The key technical novelty lies in the incorporation of visual similarity to explicitly capture dependencies between consecutive frames. We propose a dynamic programming based algorithm to efficiently evaluate all of the possible alignments and weight their importance by the consistency with the visual similarity. We further extend ECTC to incorporate frame-level semi-supervision in a unified framework, which significantly reduce the space of possible alignments. We verify the effectiveness of this framework with two applications: activity segmentation and action detection. We demonstrate that our model is able to outperform fully supervised baselines with only weak supervision, and our model achieves comparable results to state-of-the-art fully supervised models with less than 1\% of supervision.

\noindent{\bf Acknowledgement.} This work was supported by a grant from the Stanford AI Lab-Toyota Center for Artificial Intelligence Research.

\bibliographystyle{splncs03}
\bibliography{egbib}
\end{document}